Article type: Original Research

# Deep Learning Estimation of Sex, Age, Height, and Weight from CT-derived Digitally Reconstructed Radiographs

## Authors

Tomohiro Kikuchi, MD, PhD, MPH 1,2,3*, Kohei Yamamoto, MMSc 1, Yukihiro Nomura, PhD 3,4, Yosuke Yamagishi, MD 5, Takeharu Yoshikawa, MD, PhD 3, Toshiaki Akashi, MD, PhD 6, Jun Kamohara, MD 1, Hiroyuki Fujii, MD, PhD 1, Harushi Mori, MD, PhD 1

*Affiliations*

1 Department of Radiology, Jichi Medical University, 3311-1 Yakushiji, Shimotsuke-shi, Tochigi, 329-0498, Japan.
2 Data Science Center, Jichi Medical University, 3311-1 Yakushiji, Shimotsuke-shi, Tochigi, 329-0498, Japan.
3 Department of Computational Diagnostic Radiology and Preventive Medicine, The University of Tokyo Hospital, 7-3-1 Hongo, Bunkyo-ku, Tokyo, 113-8655, Japan.
4 Center for Frontier Medical Engineering, Chiba University, 1-33 Yayoicho, Inage-Ku, Chiba, Japan.
5 Division of Radiology and Biomedical Engineering, Graduate School of Medicine, The University of Tokyo, 7-3-1 Hongo, Bunkyo-Ku, Tokyo, 113-8655, Japan
6 Department of Diagnostic Radiology, Tokyo Metropolitan Institute for Geriatrics and Gerontology, 35-2 Sakae-cho, Itabashi-ku Tokyo 173-0015, Japan
Institution From Which the Work Originated:
Jichi Medical University, 3311-1 Yakushiji, Shimotsuke, Tochigi, 329-0498, Japan

## Abstract

**Purpose:** To develop and validate a deep learning ensemble for estimating adult sex, age, height, and weight from coronal digitally reconstructed radiographs (DRRs) generated from diagnostic CT.
**Materials and Methods:** This retrospective study included 128,621 CT examinations from 80,004 adults at nine institutions in Japan. Three multitask models—ConvNeXt-Base, ViT-Base/16, and MaxViT-Base—were fine-tuned using coronal DRRs and combined by weighted averaging. Data were split by institution into training (114,147 examinations; seven institutions), tuning (4,305; one institution), and test (10,169; one institution) sets; generalizability was assessed on two non-Japanese datasets. Accuracy and mean absolute error (MAE) were used to evaluate sex classification and age, height, and weight regression, respectively. Body surface area (BSA)–corrected heart and liver volume trends were compared using true versus estimated height and weight.
**Results:** In the test set (median age, 69.9 years; 4,899 of 10,169 [48.2%] male), overall sex-classification accuracy was 0.997 (95% CI, 0.996–0.998), and MAEs were 3.57 years (3.51–3.63), 2.59 cm (2.54–2.64), and 3.40 kg (3.34–3.47) for age, height, and weight, respectively. In examinations covering the chest through pelvis, accuracy was 1.000, and MAEs were 3.15 years, 2.28 cm, and 3.18 kg, respectively. BSA calculated from estimated values reproduced age-related heart and liver volume trends obtained using true values. On non-Japanese datasets, height error increased but was reduced by continued fine-tuning.
**Conclusion:** The ensemble estimated adult sex, age, height, and weight from CT-derived DRRs, with generally lower errors in examinations with broader anatomical coverage.

## Introduction

Large-scale, multi-institutional medical imaging databases are increasingly available for artificial intelligence development and epidemiological research (1,2). For the secondary use of such data, patient sex, age, height, and weight are fundamental metadata. These attributes support several applications: weight informs radiation dose optimization, such as the size-specific dose estimate (3); body size normalizes organ volumes and CT-based body composition (4–6); and sex and age stratify many epidemiological and subgroup analyses. In retrospective, shared, and anonymized data, however, these attributes are frequently missing or inconsistent.

Diagnostic CT images carry visual cues to these attributes, including skeletal morphology, body habitus, fat and muscle distribution, age-related degenerative changes, and sex differences. Several studies have used deep learning to estimate demographic and anthropometric attributes from two-dimensional projection images: sex and age can be estimated with high accuracy from chest radiographs (7–9), and weight and height have been estimated from CT scout/localizer images (10,11). Age has also been estimated directly from diagnostic CT (12,13).

Most prior reports, however, have important constraints: they address a single attribute, use a relatively small or single-institution sample, focus on a specific anatomical region, or rely on scout/localizer images. Few studies have estimated multiple attributes in adults using large-scale, multi-institutional data with institution-level external testing. Moreover, scout/localizer images are not stored or shared uniformly and are not always registered in public databases. These limitations motivate estimating these attributes from the voxel data of diagnostic three-dimensional CT itself. Because using the three-dimensional volume directly as input is computationally demanding (14,15), we instead generated a two-dimensional projection, a digitally reconstructed radiograph (DRR) (16), which is well suited to large-scale application.

The purpose of this study was to develop and validate a deep learning ensemble for estimating adult sex, age, height, and weight from coronal, real-world–scale–preserving DRRs generated from diagnostic three-dimensional CT. The models were trained on large-scale, multi-institutional data and externally validated on institutions not used for training, with additional external validation and continued fine-tuning on international datasets. To support reuse, the trained model weights and inference code are publicly available.

## Materials and Methods

This study was conducted in accordance with the principles of the Declaration of Helsinki and was approved by the Institutional Review Board of the participating institution (approval number: J24-017). The requirement for written informed consent was waived because of the retrospective nature of the study. This study was reported in accordance with the Checklist for Artificial Intelligence in Medical Imaging (CLAIM) guidelines (17). An overview of the study, including model training and inference, is shown in Figure 1.

Figure 1. Study overview

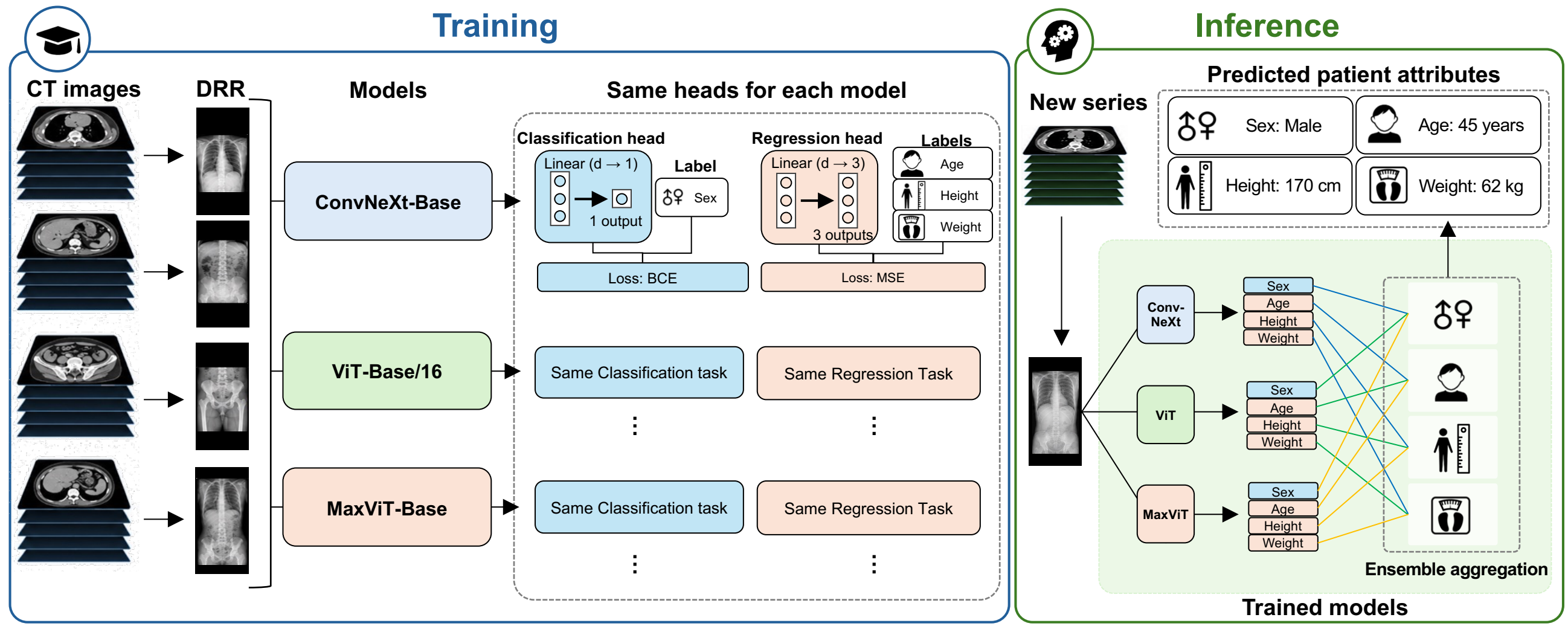

**During training, a real-world–scale–preserving coronal DRR was generated from each eligible CT series and used to fine-tune three multitask models: ConvNeXt-Base, ViT-Base/16, and MaxViT-Base. ConvNeXt-Base and ViT-Base/16 were initialized with DINOv3-pretrained weights, whereas MaxViT-Base was initialized with ImageNet-1K-pretrained weights. Each model comprised an image encoder followed by a classification head for sex, trained using BCE loss, and a regression head for age, height, and weight, trained using MSE loss. At inference, the same preprocessing was applied to a new CT series, and the outputs of the three models were combined by weighted averaging using weights optimized on the tuning set with the Nelder–Mead method. BCE = binary cross-entropy, CT = computed tomography, DRR = digitally reconstructed radiograph, MSE = mean squared error.**

*Dataset*

We used CT images acquired at university hospitals across Japan between January 1 and December 31, 2024, and provided through the Japan Medical Image Database (J-MID) (18), a multi-institutional consortium that collects clinical imaging data for research use. Data are de-identified by a prescribed procedure before provision.

Case selection is summarized in Figure 2. The inclusion criteria were the availability of sex, age, height, and weight in the DICOM metadata and an age of 18 years or older at the time of examination. Examinations were excluded if they contained predefined out-of-range values—age greater than 100 years, height less than 100 cm or greater than 210 cm, or weight less than 30 kg or greater than 200 kg—or if no axial CT series covered at least three vertebrae. Vertebral coverage was determined using vertebral labels generated by TotalSegmentator (v2.10.0) (19), with a vertebral label considered present when its segmented volume was at least 100 mm³. By these criteria, 136,272 examinations from nine institutions were eligible. No a priori sample-size calculation was performed; all eligible examinations were included.

Data were split by institution. Institutions with very small numbers of eligible examinations were retained in the training set; among the remaining institutions, the institution with the median number of eligible examinations was assigned to the test set, the institution with the next smaller number was assigned to the tuning set, and the remaining seven institutions formed the training set. The training, tuning, and test sets were therefore institutionally disjoint, with the test set serving as an institution-level external validation set. Repeat examinations of the same patient were excluded from both the tuning and test sets, leaving one examination per patient, whereas repeat examinations were retained in the training set. All eligible series were used for training, whereas one randomly selected series per examination was used for tuning and testing; evaluation was thus performed at the examination level. Imaging manufacturer, slice thickness, and the proportion of noncontrast examinations are provided in Appendix S1.

Figure 2. Flowchart of case selection.

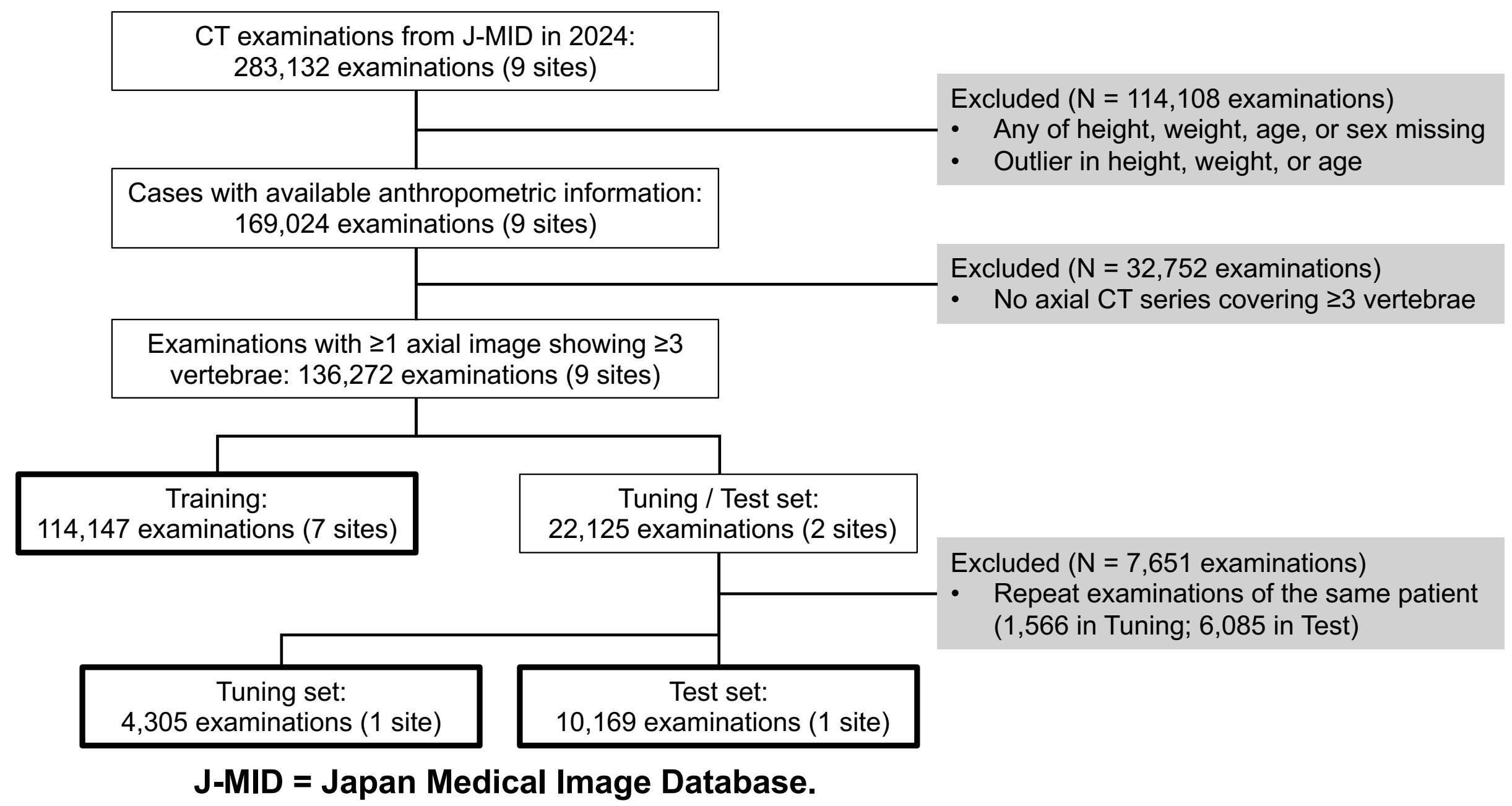

**J-MID = Japan Medical Image Database.**

For cross-domain evaluation, two public datasets were used: ENHANCE.PET (20), comprising CT acquired as part of whole-/total-body FDG-PET/CT from German and Italian centers, and RATIC (21), comprising abdominal trauma CT from institutions in multiple countries. Eligibility required only that the reference attributes be available; height and weight were not recorded in RATIC and were therefore not evaluated. The J-MID out-of-range and vertebral-coverage criteria were not applied.

*DRR Generation and Preprocessing*

From each eligible CT series, a single coronal parallel-projection DRR was generated. CT attenuation values were clipped to [−1000, 3000] Hounsfield units and converted to a relative linear attenuation coefficient map ($\mu = \mu_{_water} \times [\text{value}/1000 + 1]$, $\mu_{_water} = 0.02\ \text{mm}^{-1}$). The volume was resampled to isotropic voxels using linear interpolation and reoriented to a consistent anatomical orientation. A line integral was then computed along the anteroposterior direction to generate a coronal projection according to the Beer–Lambert law, with a fixed attenuation scaling factor of 0.005 applied to reduce intensity saturation. The resulting projection was displayed with inverted contrast and normalized to [0, 1] using min–max normalization. Parallel projection preserved the real-world scale derived from the voxel spacing. The DRR was resampled to 1.5 mm/pixel and centered on a 640 × 320-pixel canvas, with center cropping or zero padding as needed. The spatial resolution and canvas size were selected based on preliminary experiments described in Appendix S2.

*Model Architecture and Training Details*

Three deep learning models with different image encoders were developed: ConvNeXt-Base (22), ViT-Base/16 (23), and MaxViT-Base (24). ConvNeXt-Base and ViT-Base/16 were initialized with DINOv3-pretrained weights (25), whereas MaxViT-Base was initialized with ImageNet-1K-pretrained weights. Each image encoder was connected to a multitask prediction module consisting of a classification head for sex and a regression head for age, height, and body weight. The software environment was Python 3.11.14, torch 2.9.1+cu128, timm 1.0.27, and numpy 2.3.4. All models were fine-tuned end-to-end with AdamW (base learning rate, $5.0 \times 10^{-5}$ with cosine decay to a minimum of $1.0 \times 10^{-6}$ after three warmup epochs; weight decay, 0.01; $\beta_1 = 0.9$, $\beta_2 = 0.999$) for a maximum of 50 epochs with early stopping after five epochs without improvement in the tuning loss, using automatic mixed precision in the Brain Floating Point 16-bit format. The batch size was 64, except for MaxViT-Base, which used 32 because of memory constraints.

Reference labels were obtained from the DICOM metadata stored in J-MID. Because age, height, and weight differed in scale and variance, each regression target was standardized using the mean and standard deviation of the training set and back-transformed to its original scale at inference. The standardized target values were clipped to ±3 standard deviations. For each model, the training objective was the unweighted mean of the binary cross-entropy loss for sex classification and the three mean squared error losses for age, height, and weight regression.

Training was performed using one NVIDIA RTX PRO 6000 Blackwell GPU. Data augmentation included random translation, rotation, shear, horizontal and vertical flipping, Gaussian noise, and coarse dropout using random rectangular masking.

At inference, the sex probabilities and regression outputs from the three models were combined by weighted averaging. A single set of weights shared across the four prediction targets, constrained to be non-negative and to sum to one, was optimized on the tuning set using the Nelder–Mead method to minimize the combined multitask tuning loss and was subsequently fixed for evaluation on the test set; the resulting weights were 0.3954 for ConvNeXt-Base, 0.3103 for ViT-Base/16, and 0.2943 for MaxViT-Base. Sex was classified using a probability threshold of 0.5.

*Analysis and Evaluation Metrics*

Accuracy was used for sex classification, and mean absolute error (MAE) was used for age, height, and weight regression. Ninety-five percent confidence intervals were calculated using nonparametric bootstrap resampling at the patient level with 2,000 resamples and the percentile method.

To examine how performance depended on the imaged anatomical range, each examination was assigned to one mutually exclusive coverage region, and metrics were reported by region. Region membership was determined from the bone labels present: Chest was considered present when both T1 and T12 were included, Abdomen when both L1 and the iliac bone were included, and Pelvis when both L3 and the femur were included. According to the combination of basic regions present, each examination was assigned to a mutually exclusive coverage region: all three regions, Torso; Chest and Abdomen, Chest+Abdomen; Abdomen and Pelvis, Abdominopelvic; a single region only, the corresponding Chest, Abdomen, or Pelvis; and any combination not matching these, Other. Each examination thereby mapped to exactly one region, and region-specific sample sizes summed to the test total.

For Torso examinations, predicted and true values were visualized in scatter plots, and representative best, median, and worst cases were shown, with the case-selection criterion described in Appendix S3.

As a proof of concept, we simulated a database in which sex and age were available but height and weight were missing and imputed by the model, using Torso examinations in the test set. For each CT examination, heart and liver volumes were derived from TotalSegmentator (v2.10.0) segmentations and corrected by body surface area (BSA) calculated using the Du Bois formula from either true height and weight (true-BSA) or model-estimated height and weight (predicted-BSA). For each organ, BSA-corrected volume was modeled against true age using natural cubic spline regression with 4 degrees of freedom and pointwise 95% confidence intervals, stratified by true sex. The true-BSA and predicted-BSA curves were overlaid and compared qualitatively; a curve scaled by the median BSA, representing no individual body-size correction, was also included.
All cross-domain CT volumes underwent the same DRR generation and preprocessing as the J-MID data. In ENHANCE.PET, three strategies were compared: direct application of the original J-MID ensemble without additional training; continued fine-tuning of all three models from the J-MID checkpoints; and fine-tuning of the same three architectures from their original generic pretrained weights, without loading the J-MID-trained checkpoints. The latter two used the same AutoPET examinations for fine-tuning and the same Careggi examinations for tuning and ensemble-weight optimization. All three were evaluated on the Leipzig external test set. In RATIC, the original J-MID ensemble was applied without additional training.

## Results

The characteristics of the datasets are shown in Table 1. The distributions of sex, age, height, and weight were broadly similar across the J-MID splits. Median height was approximately 10 cm greater in the ENHANCE.PET cohorts than in the J-MID splits, and median age in RATIC was 44.0 years.

Table 1. Characteristics of the datasets

| Dataset | Male, n (%) | Age (years) | Height (cm) | Weight (kg) |
|---|---|---|---|---|
| **J-MID (Japan)** | | | | |
| **Training (114,147 examinations / 65,530 patients)** | 62,733 (55.0) | 70.0 [58.0, 77.0] | 162.0 [155.0, 168.0] | 58.0 [50.0, 67.0] |
| **Tuning (4,305 examinations / 4,305 patients)** | 2,106 (48.9) | 70.0 [59.0, 77.0] | 160.1 [153.6, 166.8] | 58.3 [50.6, 67.0] |
| **Test (10,169 examinations / 10,169 patients)** | 4,899 (48.2) | 69.9 [57.0, 77.0] | 160.0 [153.2, 167.0] | 59.0 [50.7, 68.2] |
| **Cross-domain datasets** | | | | |
| **ENHANCE.PET(AutoPET) (867 patients)** | 483 (55.7) | 61.0 [50.0, 72.0] | 172.0 [165.0, 179.0] | 78.0 [67.0, 88.0] |
| **ENHANCE.PET(Careggi), (199 patients)** | 126 (63.3) | 73.0 [64.5, 78.0] | 170.0 [165.0, 175.0] | 71.0 [61.5, 80.0] |
| **ENHANCE.PET (Leipzig), (381 patients)** | 270 (70.9) | 66.0 [58.3, 73.1] | 172.0 [165.0, 178.0] | 75.0 [65.0, 86.0] |
| **RATIC (2,994 patients)** | 2,101 (70.2) | 44.0 [29.0, 63.0] | — | — |

**Values are median [interquartile range] unless otherwise indicated. Because the training set retained repeat examinations of the same patient, its male counts and medians are reported at the examination level; the tuning and test sets included one examination per patient.**

Performance of the three individual models and the resulting weighted ensemble on the tuning set is shown in Table 2. The three individual models performed similarly, and after optimization of the ensemble weights, the ensemble yielded a sex-classification accuracy of 0.998 and MAEs of 3.68 years, 2.50 cm, and 3.39 kg on this set.

Performance of the ensemble applied without additional training is shown in Table 3. Within the J-MID test set, performance was generally best in Torso examinations and tended to decline as anatomical coverage narrowed, with the largest differences observed for age estimation; in Torso examinations, sex-classification accuracy was 1.000, and MAEs were 3.15 years for age, 2.28 cm for height, and 3.18 kg for weight. In the cross-domain datasets, height MAE ranged from 4.97 to 6.35 cm across the three ENHANCE.PET cohorts, and the age MAE in RATIC was 5.09 years.

Table 2. Performance of the individual models and the weighted ensemble on the tuning set.

| Model | Sex classification accuracy | Age MAE (years) | Height MAE (cm) | Weight MAE (kg) |
|---|---|---|---|---|
| **ConvNeXt-Base** | 0.995 (0.993–0.997) | 3.96 (3.87–4.06) | 2.58 (2.52–2.66) | 3.49 (3.38–3.60) |
| **ViT-Base/16** | 0.997 (0.996–0.999) | 3.92 (3.83–4.01) | 2.68 (2.61–2.75) | 3.58 (3.48–3.69) |
| **MaxViT-Base** | 0.997 (0.995–0.999) | 3.87 (3.78–3.97) | 2.70 (2.63–2.77) | 3.65 (3.54–3.75) |
| **Weighted ensemble** | 0.998 (0.997–0.999) | 3.68 (3.59–3.77) | 2.50 (2.44–2.58) | 3.39 (3.29–3.50) |

**MAE = mean absolute error. Values in parentheses are 95% confidence intervals, computed by nonparametric bootstrap with patient-level resampling (2,000 resamples, percentile method). The ensemble weights were optimized on this set.**

Table 3. Performance of the weighted ensemble by anatomical region and on cross-domain datasets.

| | Number of examinations | Sex classification accuracy | Age MAE (years) | Height MAE (cm) | Weight MAE (kg) |
|---|---|---|---|---|---|
| **J-MID test (overall)** | 10,169 | 0.997 (0.996–0.998) | 3.57 (3.51–3.63) | 2.59 (2.54–2.64) | 3.40 (3.34–3.47) |
| **Torso** | 4,232 | 1.000 (0.999–1.000) | 3.15 (3.07–3.22) | 2.28 (2.21–2.34) | 3.18 (3.07–3.30) |
| **Chest** | 2,922 | 0.998 (0.996–0.999) | 3.56 (3.45–3.67) | 2.67 (2.60–2.75) | 3.67 (3.54–3.81) |
| **Chest+Abdomen** | 469 | 0.998 (0.994–1.000) | 3.37 (3.14–3.60) | 2.56 (2.36–2.77) | 2.99 (2.77–3.24) |
| **Abdomen** | 827 | 0.992 (0.984–0.998) | 3.92 (3.74–4.13) | 3.06 (2.90–3.23) | 3.28 (3.07–3.50) |
| **Abdominopelvic** | 1,226 | 0.998 (0.996–1.000) | 4.01 (3.83–4.18) | 2.58 (2.46–2.70) | 3.16 (2.97–3.36) |
| **Pelvis** | 139 | 1.000 (1.000–1.000) | 5.07 (4.45–5.74) | 2.81 (2.34–3.30) | 3.70 (3.14–4.29) |
| **Other** | 354 | 0.969 (0.952–0.986) | 6.01 (5.57–6.50) | 4.55 (4.23–4.89) | 5.40 (4.95–5.89) |
| **ENHANCE.PET(AutoPET)** | 867 | 1.000 (1.000–1.000) | 4.34 (4.12–4.55) | 5.98 (5.70–6.31) | 4.55 (4.08–5.05) |
| **ENHANCE.PET (Careggi)** | 199 | 1.000 (1.000–1.000) | 4.58 (4.14–5.06) | 6.35 (5.67–7.00) | 3.32 (2.78–3.95) |
| **ENHANCE.PET (Leipzig)** | 381 | 0.992 (0.982–1.000) | 3.76 (3.46–4.07) | 4.97 (4.58–5.40) | 3.63 (3.24–4.08) |
| **RATIC** | 2,994 | 0.989 (0.985–0.992) | 5.09 (4.93–5.25) | — | — |

**MAE = mean absolute error. Values in parentheses are 95% confidence intervals, computed by nonparametric bootstrap with patient-level resampling (2,000 resamples, percentile method).**

For Torso examinations, which constituted the largest anatomical-coverage group in the test set, scatter plots of predicted versus true age, height, and weight are shown in Figure 3. Predictions were distributed near the line of identity for all three attributes; however, weight tended to be underestimated above approximately 100 kg. Scatter plots for the other coverage regions are shown in Appendix S4.

Figure 3. Scatter plots of predicted versus true age, height, and weight in Torso cases.

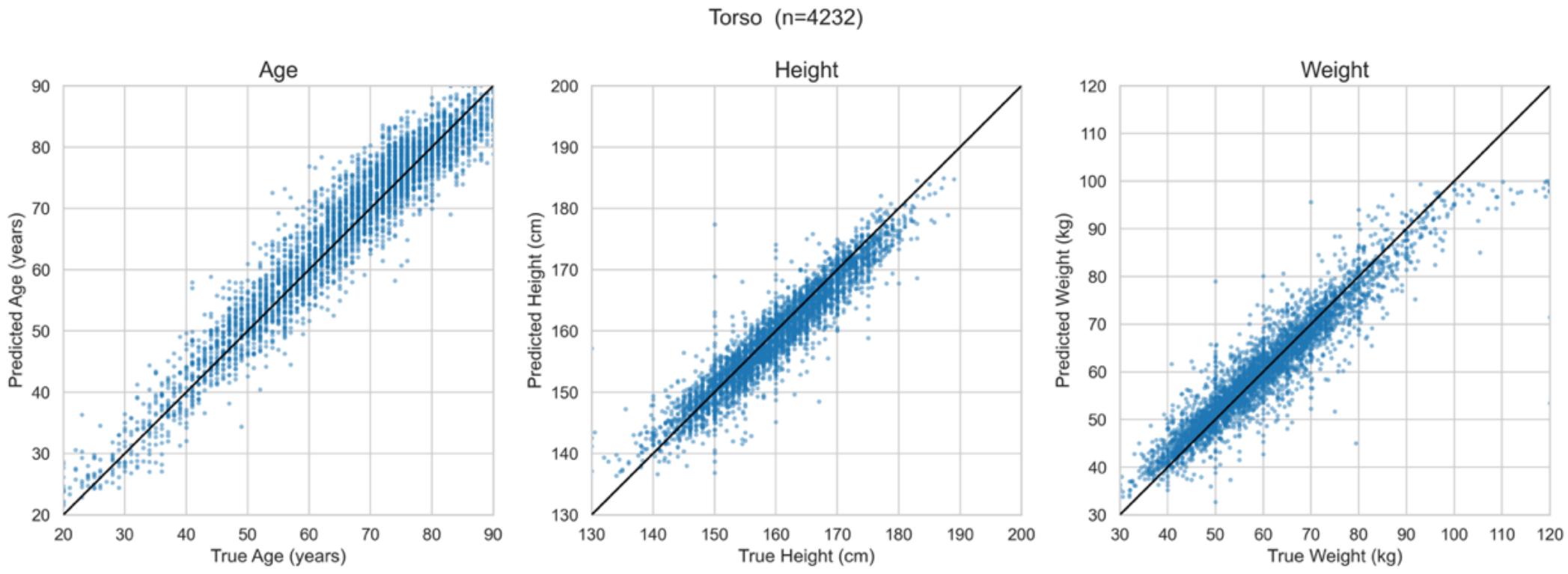

**The horizontal axis shows the true value, and the vertical axis shows the value predicted by the model. The solid line is the line of identity (perfect match), and each point represents one Torso examination. Predictions cluster near the line of identity for all three attributes. n = number of Torso examinations.**

Representative best, median, and worst Torso cases are shown in Figure 4. Representative cases for the Chest, Abdominopelvic, and Other regions are shown in Appendix S5.

Figure 4. Representative inference results in Torso cases.

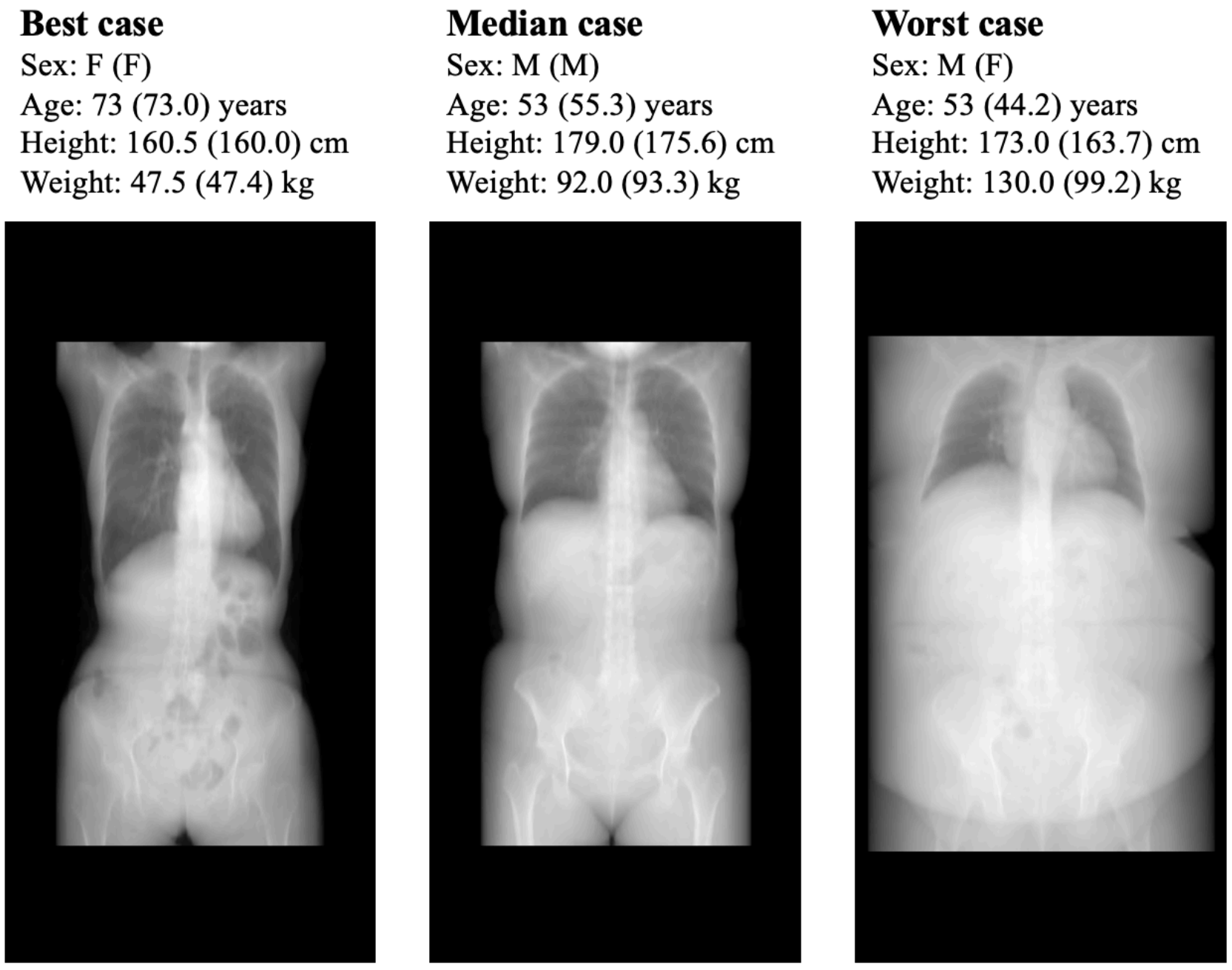

**The best, median, and worst inference results are shown. In each panel, sex, age, height, and weight are given as the true value and the model-predicted value in parentheses, overlaid on the input coronal digitally reconstructed radiograph. M = male, F = female.**

Figure 5 shows heart and liver volumes in Torso test examinations together with the fitted regression curves. At the population level, the predicted-BSA and true-BSA curves were closely superimposed, and their 95% confidence intervals overlapped throughout, whereas the uncorrected curves deviated from the individually BSA-corrected curves over some age ranges.

Figure 5. Body surface area (BSA)–corrected heart and liver volumes versus age.

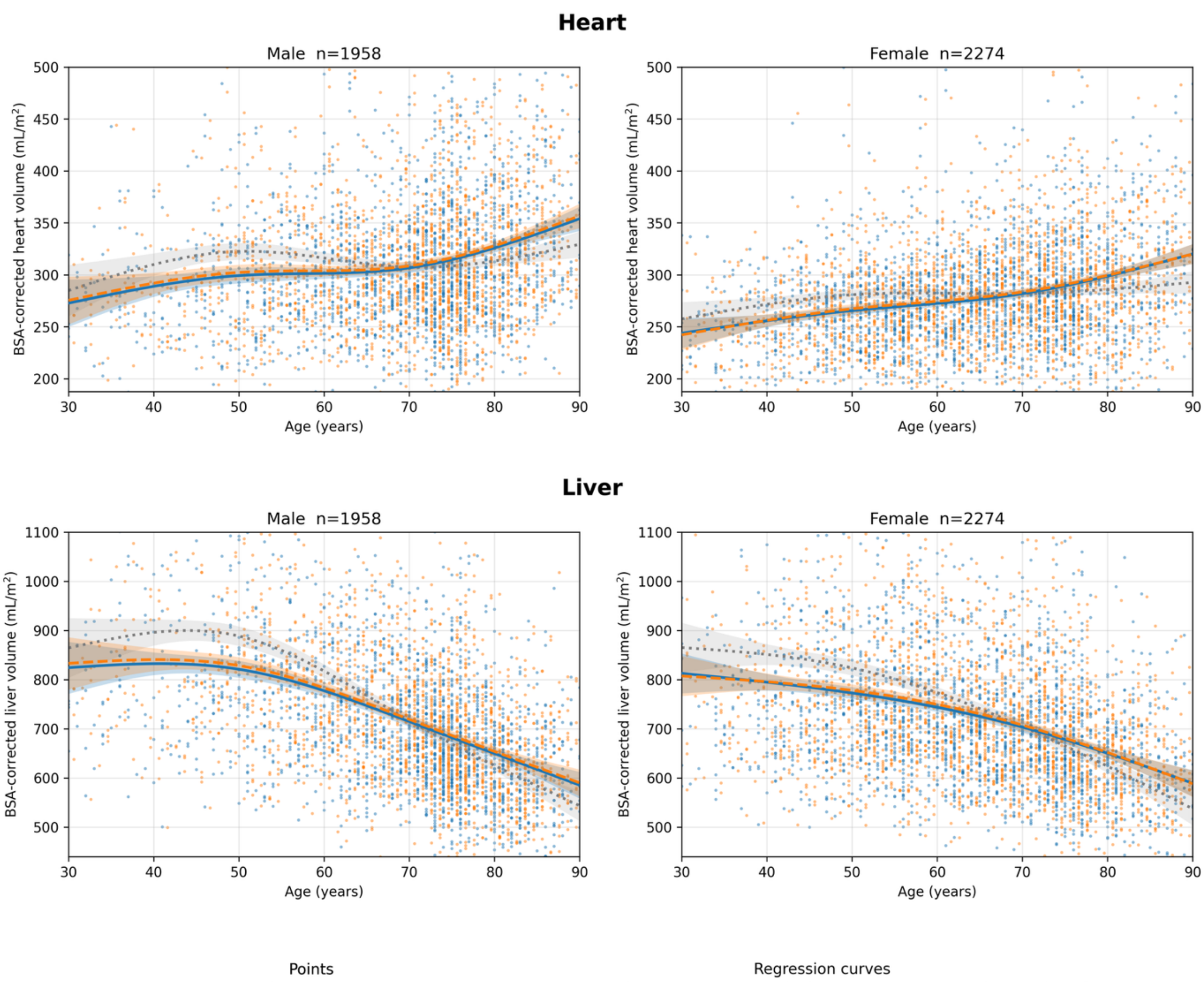

**The top and bottom rows show the heart and liver, and the left and right columns show males and females. Three curves are shown: BSA from true height and weight (true-BSA), BSA from model-estimated height and weight (predicted-BSA), and a curve scaled by the median BSA, representing no individual body-size correction (uncorrected). Shaded bands of the same color as each curve denote the pointwise 95% confidence intervals.**
**BSA = body surface area, n = number of examinations, predicted-BSA = body surface area calculated from model-estimated height and weight, true-BSA = body surface area calculated from true height and weight.**

The effect of initialization from the J-MID checkpoints is shown in Table 4. In the Leipzig external test set, continued fine-tuning reduced the height MAE from 4.97 to 3.47 cm and the weight MAE from 3.63 to 3.55 kg, whereas the age MAE was 3.91 years compared with 3.76 years without additional training. Fine-tuning the same architectures without initialization from the J-MID checkpoints yielded higher errors for all four attributes (accuracy, 0.958; MAEs, 5.74 years, 4.10 cm, and 4.83 kg).

*Table 4. Performance on cross-domain datasets.*

| | Sex classification accuracy | Age MAE (years) | Height MAE (cm) | Weight MAE (kg) |
|---|---|---|---|---|
| **ENHANCE.PET, Leipzig external test (Germany; n = 381)** | | | | |
| Base ensemble (no additional training) | **0.992 (0.982–1.000)** | **3.76 (3.46–4.07)** | 4.97 (4.58–5.40) | 3.63 (3.24–4.08) |
| Continued fine-tuning from our checkpoints | **0.992 (0.982–1.000)** | 3.91 (3.64–4.20) | **3.47 (3.17–3.78)** | **3.55 (3.25–3.89)** |
| Fine-tuning from generic pretrained initialization | 0.958 (0.937–0.976) | 5.74 (5.30–6.19) | 4.10 (3.77–4.43) | 4.83 (4.44–5.21) |

**MAE = mean absolute error. Values in parentheses are 95% confidence intervals, computed by nonparametric bootstrap with patient-level resampling (2,000 resamples, percentile method). For the Leipzig external test set, the best value in each column is shown in bold.**

## Discussion

In this study, a weighted ensemble estimated adult sex, age, height, and weight from real-world–scale–preserving DRRs generated from diagnostic three-dimensional CT. On the overall external test set, sex-classification accuracy was 0.997, and MAEs were 3.57 years, 2.59 cm, and 3.40 kg. This study integrates four-attribute prediction, large-scale multi-institutional training, and institution-level external testing, with public release of the model weights and inference code.

Sex prediction approached a ceiling (Torso accuracy, 1.000), consistent with prior chest-radiograph studies reporting an area under the receiver operating characteristic curve of approximately 0.94 or accuracy exceeding 0.99 (8,26). This high performance likely reflects prominent sex-related differences in pelvic morphology, thoracic cage shape, shoulder width, and soft-tissue distribution captured by the coronal projection (26,27). Age estimation in Torso examinations (MAE, 3.15 years) yielded a lower error than that reported in a previous study using chest–abdomen CT in a clinical adult population (external-test MAE, 6.50 years) (12) and approached the performance of a multi-institutional chest radiography model developed in healthy individuals (external-test MAE, 3.0 years) (9). This comparison is notable because the present cohort was older and clinically broader. Age estimation was the most coverage-dependent, with the MAE increasing to 5.07 years for Pelvis and 6.01 years for the Other category. Because many clinical CT examinations cover only part of the body, useful performance across partial-coverage regions may broaden practical applicability.

Height and weight were also estimated with low errors in Torso examinations (MAEs, 2.28 cm and 3.18 kg, respectively). These errors were lower than the 5.58-cm and 4.25-kg MAEs reported using CT localizers in children and young adults (11). The weight MAE was also within the range of 2.75–4.77 kg reported in previous studies using CT scout images (3,10). However, weight tended to be underestimated above approximately 100 kg. This finding may partly reflect clipping of the standardized training targets at ±3 standard deviations, together with the small number of individuals with very high body weight in the right-skewed training distribution. Improving performance at the extremes of the weight distribution remains an area for future investigation.

As a proof of concept, BSA calculated from predicted height and weight reproduced age-related heart and liver volume trends obtained using true BSA. Because body-size normalization is common in organ-volume and body-composition studies (4–6), this suggests utility for databases with incomplete anthropometric metadata. Although privacy-related inference was not an objective of this study, image-based estimation of patient attributes carries broader implications and warrants caution: similar inference has been reported for other attributes and imaging modalities, including race and, to a lesser extent, insurance status (8,28), and such capabilities merit careful consideration as these methods continue to develop.

Applied to datasets that differed from J-MID in country, acquisition context, and clinical population, the ensemble degraded unevenly rather than uniformly. Height estimation was markedly worse in all three ENHANCE.PET cohorts, whereas sex, age, and weight estimation were much less affected; in RATIC, age estimation showed the largest error. This pattern parallels the reference distributions: median height was approximately 10 cm greater in the ENHANCE.PET cohorts than in J-MID, and median age in RATIC was 44.0 years compared with 69.9 years in the J-MID test set. These findings suggest that cross-domain degradation may reflect differences in anthropometric distributions, acquisition protocols, clinical populations, or reference-label quality, although their respective contributions could not be isolated. Continued fine-tuning from the J-MID checkpoints improved height and weight estimation relative to direct transfer and outperformed fine-tuning from generic pretrained initialization across all four targets.

This study has several limitations. First, the reference labels were derived from DICOM metadata, and the timing of their acquisition and whether height and weight were measured or self-reported were unknown. Second, the training data were derived entirely from institutions in Japan, and although cross-domain analyses were performed, generalizability to broader international and ethnically diverse populations remains uncertain. Third, the training data were skewed toward older adults (median age, approximately 70 years). Because the cohort reflected a clinical rather than screening population, performance in healthy populations, younger adults, and individuals at the extremes of the anthropometric distributions remains to be established. Fourth, only a single coronal projection was used; weight estimation in particular may benefit from anteroposterior information. Although three-dimensional processing could incorporate this, it would require greater computation and larger datasets; we adopted a single two-dimensional projection for feasibility (14,15).

In conclusion, a deep learning ensemble estimated sex, age, height, and weight from real-world–scale–preserving DRRs derived from diagnostic CT, with generally lower errors in examinations with broader anatomical coverage. As a proof of concept, BSA calculated from estimated height and weight reproduced age-related organ-volume trends obtained using true values.

## Acknowledgments

The authors thank the departments of radiology that contribute to the Japan Medical Image Database (J-MID), including Juntendo University, Kyushu University, Keio University, The University of Tokyo, Okayama University, Kyoto University, Osaka University, Tokyo Medical and Dental University, Hokkaido University, Ehime University, and Tokushima University, for their contributions to the database.

## Conflicts of Interest

The authors declare no conflicts of interest.

## Funding

This research was partially supported by JSPS KAKENHI [Grant Number JP26K19038].

## Data sharing statement

Individual-level deidentified CT images and associated metadata obtained from J-MID cannot be shared by the authors because of the terms governing access to the database and applicable privacy restrictions. The ENHANCE.PET and RATIC datasets used for cross-domain evaluation are available through their respective public repositories under the applicable access and data-use conditions, as described in references 20 and 21. The trained model weights and source code for DRR generation, training, evaluation, and inference are publicly available at: https://github.com/jichi-labo/DRRBiometricsPredictor

## References

1. Littlejohns TJ, Holliday J, Gibson LM, et al. The UK Biobank imaging enhancement of 100,000 participants: rationale, data collection, management and future directions. Nat Commun 2020;11(1):2624. doi: 10.1038/s41467-020-15948-9
2. Hamamci IE, Er S, Wang C, et al. Generalist foundation models from a multimodal dataset for 3D computed tomography. Nat Biomed Eng 2026. doi: 10.1038/s41551-025-01599-y. Published February 12, 2026. Accessed June 28, 2026
3. Ichikawa S, Itadani H, Sugimori H. Deep learning-based body weight from scout images can be an alternative to actual body weight in CT radiation dose management. J Appl Clin Med Phys 2023;24(8):e14080. doi: 10.1002/acm2.14080
4. Kikuchi T, Yamamoto K, Yamagishi Y, et al. Nationwide organ volume distributions and cross-sectional age-associated differences in abdominal CT from Japan. Jpn J Radiol 2026. doi: 10.1007/s11604-026-02028-z. Published June 11, 2026. Accessed June 28, 2026
5. Wachinger C, Renger B, Späth C, Makowski MR. Body charts from CT segmentations across the adult lifespan: large-scale cross-sectional and longitudinal analyses. Radiol Artif Intell 2026;8(2):e250506. doi: 10.1148/ryai.250506
6. Blankemeier L, Yao L, Long J, et al. Skeletal muscle area on CT: determination of an optimal height scaling power and testing for mortality risk prediction. AJR Am J Roentgenol 2024;222(1):e2329889. doi: 10.2214/AJR.23.29889
7. Yi PH, Wei J, Kim TK, et al. Radiology “forensics”: determination of age and sex from chest radiographs using deep learning. Emerg Radiol 2021;28(5):949–954. doi: 10.1007/s10140-021-01953-y
8. Adleberg J, Wardeh A, Doo FX, et al. Predicting patient demographics from chest radiographs with deep learning. J Am Coll Radiol 2022;19(10):1151–1161. doi: 10.1016/j.jacr.2022.06.008
9. Mitsuyama Y, Matsumoto T, Tatekawa H, et al. Chest radiography as a biomarker of ageing: artificial intelligence-based, multi-institutional model development and validation in Japan. Lancet Healthy Longev 2023;4(9):e478–e486. doi: 10.1016/S2666-7568(23)00133-2
10. Ichikawa S, Hamada M, Sugimori H. A deep-learning method using computed tomography scout images for estimating patient body weight. Sci Rep 2021;11(1):15627. doi: 10.1038/s41598-021-95170-9
11. Demircioğlu A, Quinsten AS, Umutlu L, Forsting M, Nassenstein K, Bos D. Determining body height and weight from thoracic and abdominal CT localizers in pediatric and young adult patients using deep learning. Sci Rep 2023;13(1):19010. doi: 10.1038/s41598-023-46080-5
12. Kerber B, Hepp T, Küstner T, Gatidis S. Deep learning-based age estimation from clinical computed tomography image data of the thorax and abdomen in the adult population. PLoS One 2023;18(11):e0292993. doi: 10.1371/journal.pone.0292993

13. Farina EMJM, Matsuoka FA, Corradi G, et al. Pixel tampering: does face redaction harm medical AI performance? J Imaging Inform Med 2025. doi: 10.1007/s10278-025-01776-0. Published December 16, 2025. Accessed June 28, 2026

14. Singh SP, Wang L, Gupta S, Goli H, Padmanabhan P, Gulyás B. 3D deep learning on medical images: a review. Sensors (Basel) 2020;20(18):5097. doi: 10.3390/s20185097

15. Fu J, Yang Y, Singhrao K, et al. Deep learning approaches using 2D and 3D convolutional neural networks for generating male pelvic synthetic computed tomography from magnetic resonance imaging. Med Phys 2019;46(9):3788–3798. doi: 10.1002/mp.13672

16. Siddon RL. Fast calculation of the exact radiological path for a three-dimensional CT array. Med Phys 1985;12(2):252–255. doi: 10.1118/1.595715

17. Tejani AS, Klontzas ME, Gatti AA, et al. Checklist for artificial intelligence in medical imaging (CLAIM): 2024 update. Radiol Artif Intell 2024;6(4):e240300. doi: 10.1148/ryai.240300

18. Akashi T, Kumamaru KK, Wada A, et al. Japan-Medical Image Database (J-MID): medical big data supporting data science. Juntendo Med J 2025;71(3):166–172. doi: 10.14789/ejmj.JMJ25-0004-P

19. Wasserthal J, Breit HC, Meyer MT, et al. TotalSegmentator: robust segmentation of 104 anatomic structures in CT images. Radiol Artif Intell 2023;5(5):e230024. doi: 10.1148/ryai.230024

20. Ferrara D, Pires M, Gutschmayer S, et al. Sharing a whole-/total-body [$^{18}$F]FDG-PET/CT dataset with CT-derived segmentations: an ENHANCE.PET initiative. Sci Data 2026;13:869. doi: 10.1038/s41597-026-07218-y

21. Rudie JD, Lin HM, Ball RL, et al. The RSNA Abdominal Traumatic Injury CT (RATIC) dataset. Radiol Artif Intell 2024;6(6):e240101. doi: 10.1148/ryai.240101

22. Liu Z, Mao H, Wu CY, Feichtenhofer C, Darrell T, Xie S. A ConvNet for the 2020s. In: Proceedings of the 2022 IEEE/CVF Conference on Computer Vision and Pattern Recognition. Piscataway, NJ: IEEE; 2022:11966–11976. doi: 10.1109/CVPR52688.2022.01167

23. Dosovitskiy A, Beyer L, Kolesnikov A, et al. An image is worth 16×16 words: transformers for image recognition at scale. Presented at: International Conference on Learning Representations (ICLR); May 3–7, 2021; virtual

24. Tu Z, Talebi H, Zhang H, et al. MaxViT: multi-axis vision transformer. In: Avidan S, Brostow G, Cissé M, Farinella GM, Hassner T, eds. Computer Vision—ECCV 2022. Lecture Notes in Computer Science, vol 13684. Cham, Switzerland: Springer; 2022:459–479. doi: 10.1007/978-3-031-20053-3_27

25. Siméoni O, Vo HV, Seitzer M, et al. DINOv3. arXiv:2508.10104 [preprint]. Posted August 13, 2025. Accessed June 28, 2026

26. Li D, Lin CT, Sulam J, Yi PH. Deep learning prediction of sex on chest radiographs: a potential contributor to biased algorithms. Emerg Radiol 2022;29(2):365–370. doi: 10.1007/s10140-022-02019-3

27. Torimitsu S, Makino Y, Saitoh H, et al. Morphometric analysis of sex differences in contemporary Japanese pelves using multidetector computed tomography. Forensic Sci Int 2015;257:530.e1–530.e7. doi: 10.1016/j.forsciint.2015.10.018

28. Gichoya JW, Banerjee I, Bhimireddy AR, et al. AI recognition of patient race in medical imaging: a modelling study. Lancet Digit Health 2022;4(6):e406–e414. doi: 10.1016/S2589-7500(22)00063-2

# Appendix

## S1. Imaging characteristics of the training, tuning, and test datasets

| Attribute | Train | Tuning | Test |
|---|---|---|---|
| No. of institutions | 7 | 1 | 1 |
| No. of examinations | 114,147 | 4,305 | 10,169 |
| Manufacturer, n (%)<br>(examination-level) | Toshiba 53,169 (46.6%)<br>Canon 35,361 (31.0%)<br>Siemens 14,475 (12.7%)<br>GE 5,849 (5.1%)<br>Philips 5,293 (4.6%) | Canon 1,918 (44.6%)<br>Philips 2,387 (55.4%) | Canon 5,476 (53.8%)<br>Siemens 4,693 (46.2%) |
| No. of series | 293,356 | 4,305 | 10,169 |
| Noncontrast series, n (%) | 140,055 (47.7%) | 4,037 (93.8%) | 7,071 (69.5%) |
| Slice thickness | 5 mm (89.7%)<br>range: 2.5–10 mm | 5 mm (99.5%)<br>range: 3–5 mm | 5 mm (96.0%)<br>range: 3–10 mm |
| For tuning and testing, one series per examination was used; therefore, the numbers of series and examinations are identical. | | | |

## S2. Selection of DRR spatial resolution and input canvas size

Figure S2: Examples of digitally reconstructed radiographs generated using different spatial resolutions and input canvas sizes.

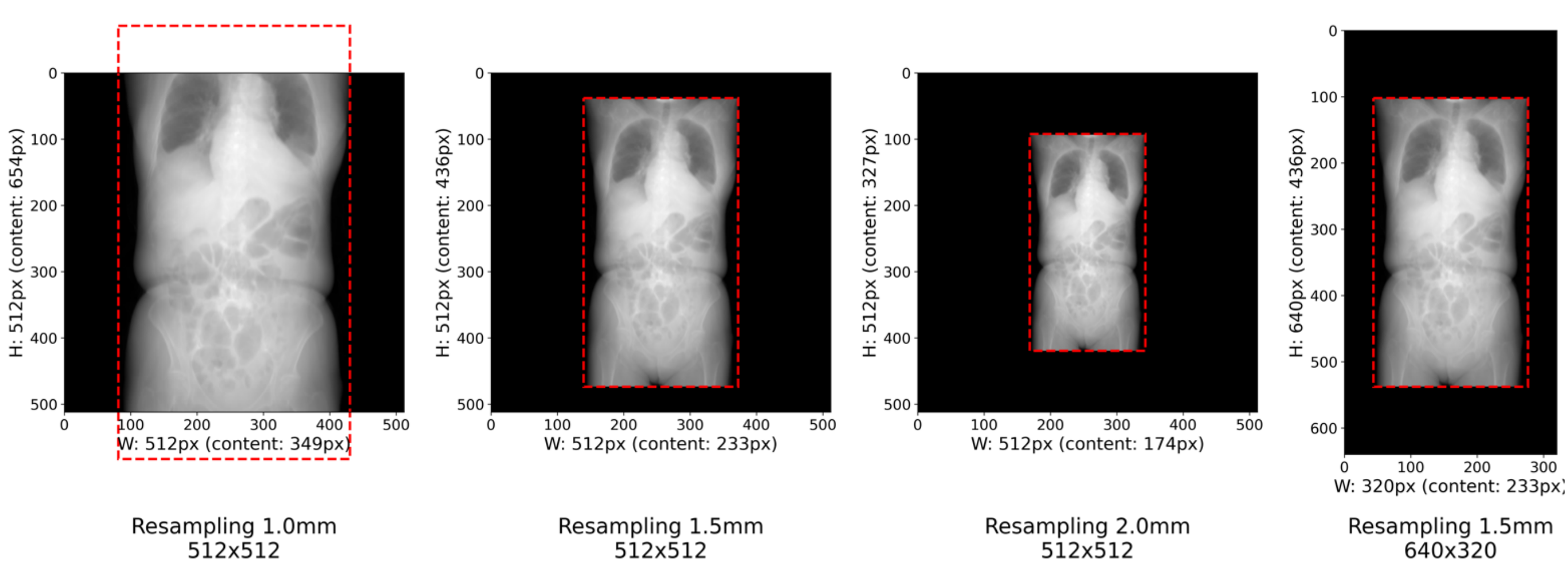

- A representative examination of an individual approximately 165 cm tall is shown. The red dashed rectangle indicates the full extent of the resampled DRR before center cropping or zero padding to the specified canvas size.

Table S2: Performance across DRR spatial resolutions and input canvas sizes in Torso examinations from the tuning set.

| Resample (Image Size) | N | Sex classification accuracy | Age MAE (years) | Height MAE (cm) | Weight MAE (kg) |
|---|---|---|---|---|---|
| **1.0 mm (512×512)** | 1,324 | **1.000 (1.000–1.000)** | 3.67 (3.50–3.84) | 2.43 (2.33–2.54) | 3.47 (3.29–3.67) |
| **1.5 mm (512×512)** | | 0.999 (0.998–1.000) | 3.61 (3.45–3.77) | 2.20 (2.11–2.30) | 3.23 (3.04–3.43) |
| **2.0 mm (512×512)** | | 0.999 (0.998–1.000) | 3.67 (3.51–3.83) | 2.33 (2.23–2.43) | **3.18 (3.00–3.38)** |
| **1.5 mm (640×320)** | | **1.000 (1.000–1.000)** | **3.48 (3.33–3.63)** | **2.17 (2.08–2.27)** | **3.18 (3.00–3.37)** |
| MAE = mean absolute error. The best value in each column is shown in bold. | | | | | |

- Preliminary experiments were performed using the ConvNeXt-Base model on 1,324 Torso examinations from the tuning set to select the DRR spatial resolution and input canvas size. Torso examinations were used because they provided broad anatomical coverage and generally showed the best prediction performance. Sex-classification accuracy was near the ceiling for all configurations. A spatial resolution of 1.5 mm/pixel with a 640 × 320-pixel canvas achieved the lowest MAEs for age and height and tied for the lowest MAE for weight; therefore, this configuration was adopted for all subsequent experiments.

# S3. Standardized composite error for case selection

The case-selection metric was a standardized composite error that integrated prediction errors for age, height, weight, and sex. For age, height, and weight, the absolute difference between the predicted and true values was divided by the standard deviation of each attribute across the entire test population; for sex, misclassification was standardized by the standard deviation of a Bernoulli distribution. To suppress the influence of extreme outliers, each standardized error was clipped at an upper bound of 3.0, and the mean of the four components was taken as the per-case composite error. Smaller values indicate closer overall agreement with the true values across the four attributes.

- Composite error =

  $1/4 \times \{ \min(|age_{pred} - age_{true}| / SD(age_{true}), 3)$

  $+ \min(|height_{pred} - height_{true}| / SD(height_{true}), 3)$

  $+ \min(|weight_{pred} - weight_{true}| / SD(weight_{true}), 3)$

  $+ \min(I(sex_{pred} \neq sex_{true}) / \sqrt{(p_sex(1 - p_{sex}))}, 3) \}$

where SD denotes the standard deviation over the test population, I(·) the indicator function, and p_sex the proportion of male examinations in the test set.
Using this metric, in each of the Chest, Abdominopelvic, and Other regions, the case with the smallest composite error was identified as the best case, the case with the largest composite error as the worst case, and the case with the median composite error as the median case for visualization. Torso cases are described in the main text.

## S4. Scatter plots for the coverage regions other than Torso.

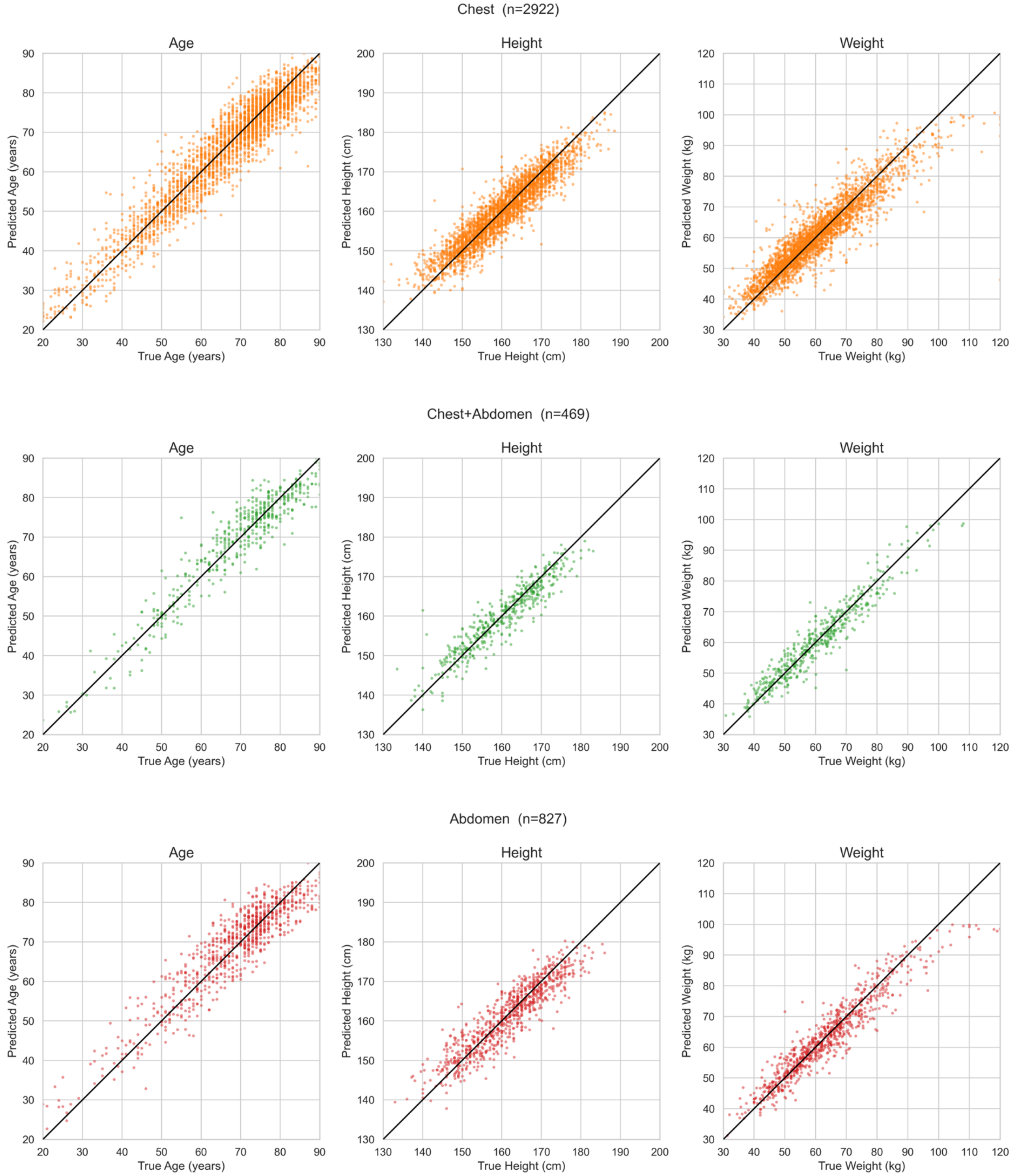

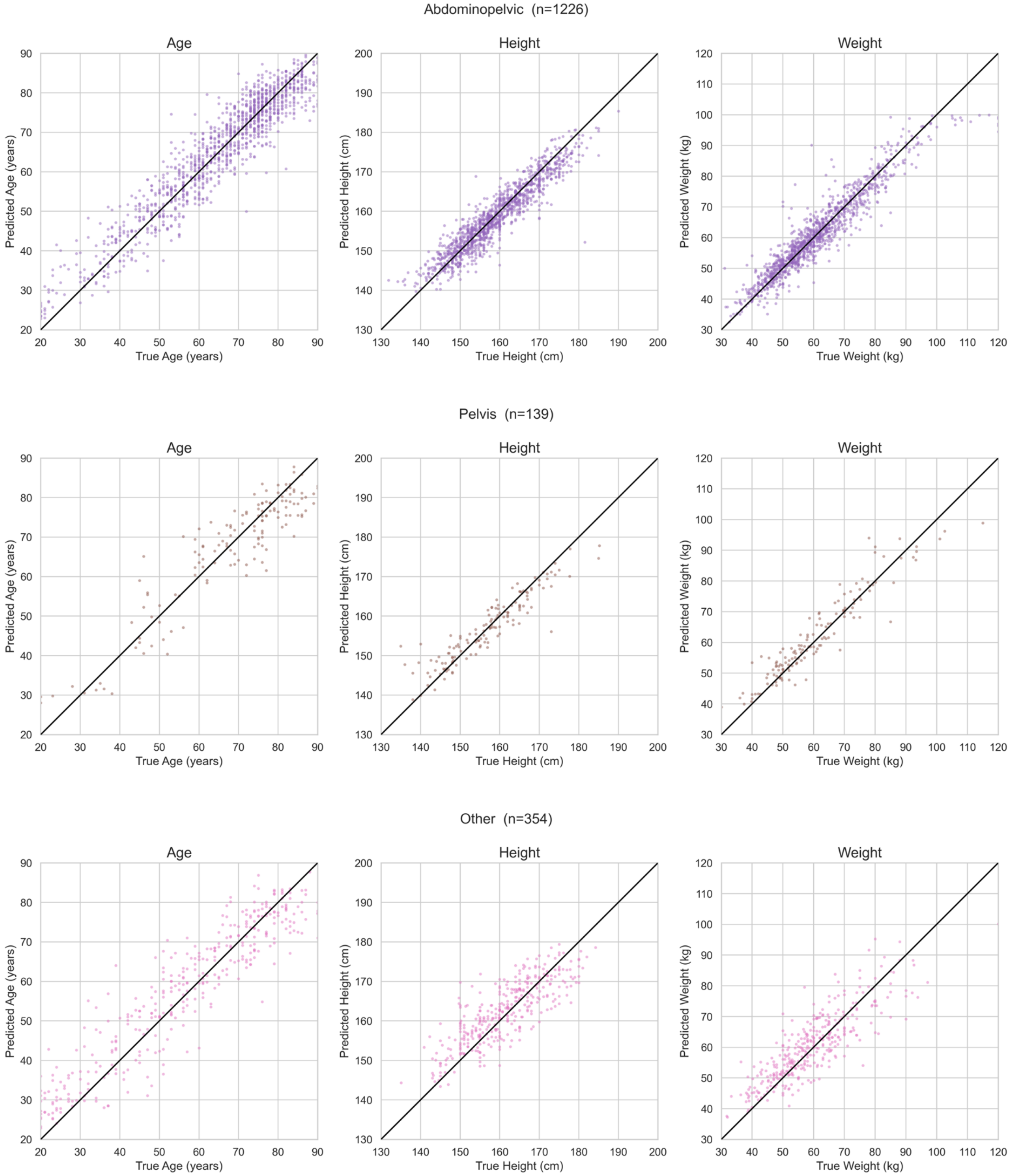

- Scatter plots of predicted versus true age, height, and weight are shown for the Chest, Chest+Abdomen, Abdomen, Abdominopelvic, Pelvis, and Other regions. The axes and line of identity are the same as those in Figure 3.

## S5. Representative inference examples for the Chest, Abdominopelvic, and Other regions.

- The best, median, and worst cases are shown, selected using the composite error defined in Appendix S3. In each panel, sex, age, height, and weight are shown as the true value, with the model-predicted value in parentheses. Torso cases are shown in Figure 4.

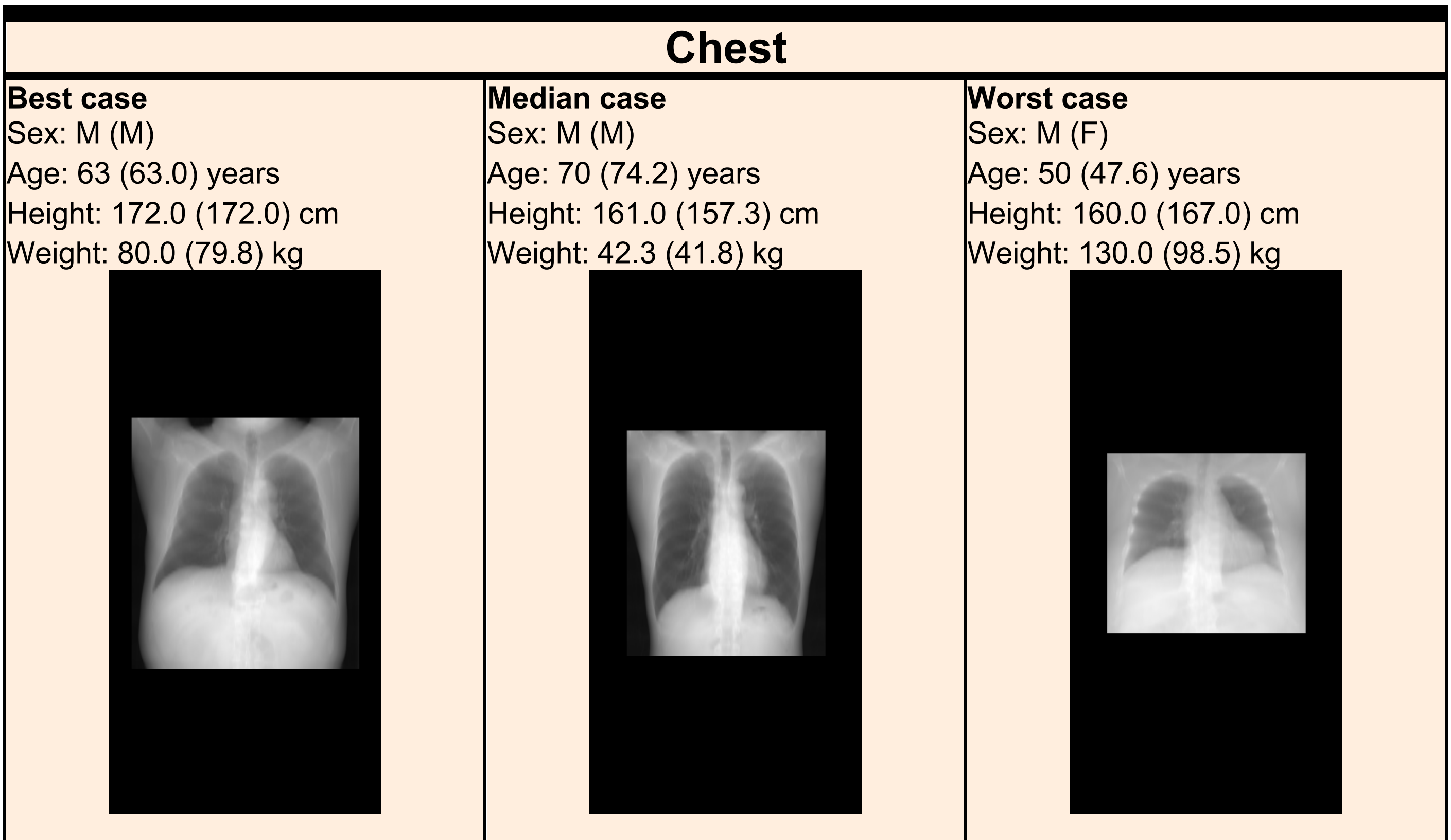

**Abdominopelvic**

**Best case**
Sex: M (M)
Age: 66 (66.9) years
Height: 162.0 (162.0) cm
Weight: 60.5 (60.5) kg

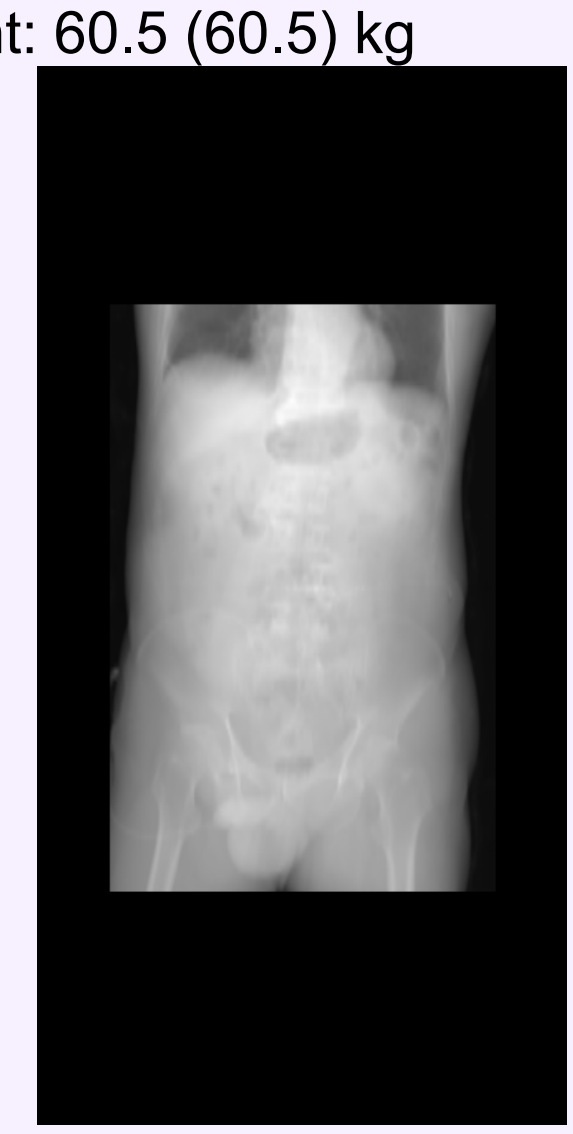

**Median case**
Sex: M (M)
Age: 60 (63.3) years
Height: 176.0 (172.7) cm
Weight: 68.0 (69.6) kg

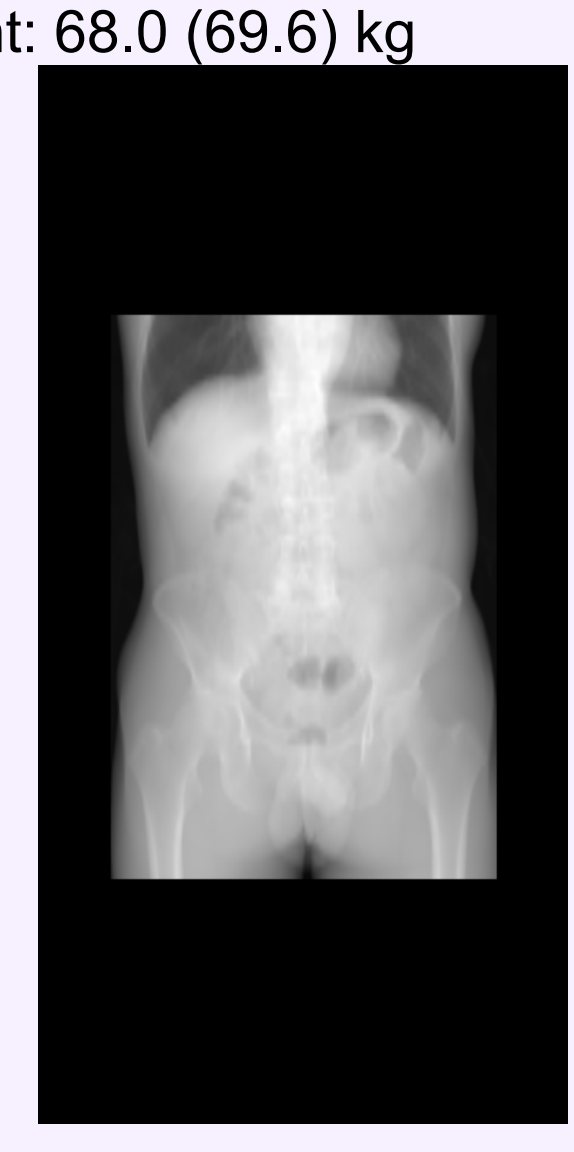

**Worst case**
Sex: M (M)
Age: 47 (58.0) years
Height: 185.0 (174.2) cm
Weight: 129.5 (98.7) kg

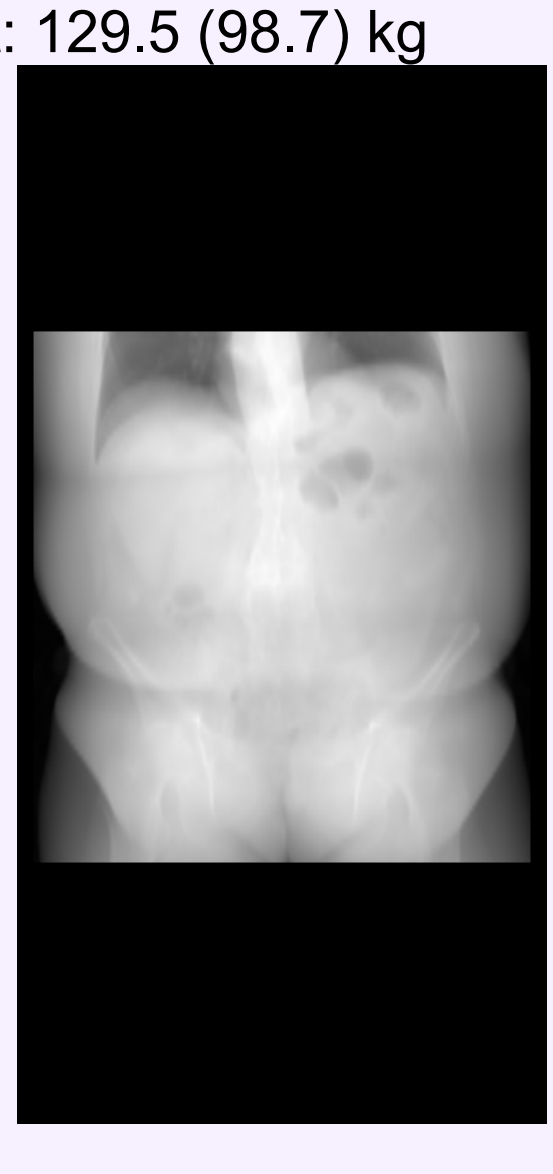

| Other | | |
|---|---|---|
| **Best case**<br>Sex: F (F)<br>Age: 81 (82.3) years<br>Height: 144.6 (145.0) cm<br>Weight: 50.6 (51.4) kg | **Median case**<br>Sex: F (F)<br>Age: 79 (77.5) years<br>Height: 160.0 (153.4) cm<br>Weight: 55.0 (49.9) kg | **Worst case**<br>Sex: F (M)<br>Age: 51 (67.7) years<br>Height: 140.9 (153.3) cm<br>Weight: 39.0 (42.5) kg |